\documentclass[conference,9pt]{IEEEtran}
\IEEEoverridecommandlockouts
\usepackage{cite}
\usepackage{amsmath,amssymb,amsfonts}
\usepackage{algorithmic}
\usepackage{graphicx}
\usepackage{textcomp}
\usepackage{xcolor}
\usepackage{booktabs}
\usepackage{xspace}
\usepackage{multirow}
\usepackage{latexsym}
\usepackage{enumitem}
\usepackage[T1]{fontenc}
\usepackage[utf8]{inputenc}
\usepackage{microtype}
\usepackage{inconsolata}
\usepackage{etoolbox}
\usepackage{hyperref}
\def\BibTeX{{\rm B\kern-.05em{\sc i\kern-.025em b}\kern-.08em T\kern-.1667em\lower.7ex\hbox{E}\kern-.125emX}}

\begin{document}

\title{UPCS: Unbiased Persona Construction for Dialogue Generation\\
}

\author{
\IEEEauthorblockN{Kuiyun Chen$^{\star}$}
\IEEEauthorblockA{
\textit{School of Software} \\
\textit{Shandong University, China} \\
Jinan, China \\
Timothybinette3453@gmail.com
}
\and
\IEEEauthorblockN{Yanbin Wei$^{\star\dagger}$\thanks{$^{\star}$ Equal contribution. $^{\dagger}$ Corresponding author.}}
\IEEEauthorblockA{
\textit{Hong Kong University of Science and Technology, Hong Kong}\\
\textit{Southern University of Science and Technology, China} \\
Hong Kong SAR, China \\
yanbin.ust@gmail.com
}
}

\maketitle

\begin{abstract}
Narrative systems, such as dialogue and storytelling systems, often utilize persona profiles to enhance personalized interactions. Existing persona profiles frequently exhibit biases, posing risks to system integrity and fairness. To address this, we introduce the UPCS framework, which categorizes character descriptions into eight dimensions, including bias mitigation strategies. Experimental results demonstrate UPCS's superiority in accuracy, diversity, bias elimination, and user satisfaction, marking a significant advancement in persona construction for reliable narrative systems.
\end{abstract}


\section{Introduction}

\label{sec:intro}
The integration of personalized dialogue systems, such as chatbots, into contemporary communication channels, aims to elevate user experiences by rendering interactions more immersive and akin to human-like exchanges \cite{Lu2023MIRACLE, Li2022ToD4IR}. A principal methodology employed to achieve this objective is the "Persona" strategy, wherein virtual characters are endowed with distinct personality traits and elaborate backstories \cite{Wang2023Does, Zhao2023Dialogue}. This technique enhances the enjoyment and appeal of conversations, thereby increasing user engagement and satisfaction. Nevertheless, while the implementation of Personas offers numerous benefits, the methods used for their construction present potential challenges, notably the risk of embedding toxic biases.  \cite{Deshpande2023Toxicity,Wan2023Are}.

Mainstream Persona construction methods can be categorized into three types, each bearing the risk of introducing toxic biases: 

\textit{1) Directly Extracting Personas}: Techniques such as \cite{Zhu2023PAED:} and \cite{delucia2024usingnaturallanguageinference} rely on existing natural dialogue data to directly extract character settings (Persona). However, the natural dialogues used as material may contain biases related to genders, races, ages, religions or other social factors \cite{10.1145/3597307}. For instance, a travel review website predominantly catering to Western audiences may exhibit a significant underrepresentation of Muslim users from the Middle East. If the data from these reviews are utilized to extract Personas, it could result in Personas that lack an understanding and respect for Middle Eastern cultures and Islamic religious practices. Consequently, such Personas might not only fail to accurately reflect the perspectives and needs of this group but could also inadvertently perpetuate stereotypes about the region and its religious practices, leading to biased and unfair service experiences for these users. 

\textit{2) Manual-defined Personas}: Approaches like \cite{Zheng2019A} and \cite{gao2023peacokpersonacommonsenseknowledge} depend on experts or users manually creating character settings. Although predefined Persona methods can mitigate bias through human oversight, they may still be influenced by the inherent biases of the creators. For instance, creators might set character attributes based on personal experiences or societal stereotypes, which can be reflected and amplified in the Personas. 

3) \textit{Automatically Generated Personas}: Leveraging the capabilities of generative models, such as large language models, which excel at handling structured and complex data (e.g. social network graphs \cite{wasserman1994social,wei2024rendering}),  methods like those mentioned by \cite{bdcc4030021} and \cite{10.1007/978-3-030-50334-5_7} aim to train generative models on extensive social media data to automatically create Personas. Compared to direct extraction from natural dialog, this approach can generate a larger number of Personas by learning the distribution of Personas within the data. However, various biases present in the data can be learned by the models and reflected in the generated Personas. Besides, the generated texts with the models are also illustrated as unreliable in many sensitive and adversarial topics.\cite{lu2023large,lu2024less}.

\begin{figure*}[htbp]
    \centering
    \includegraphics[width=\textwidth]{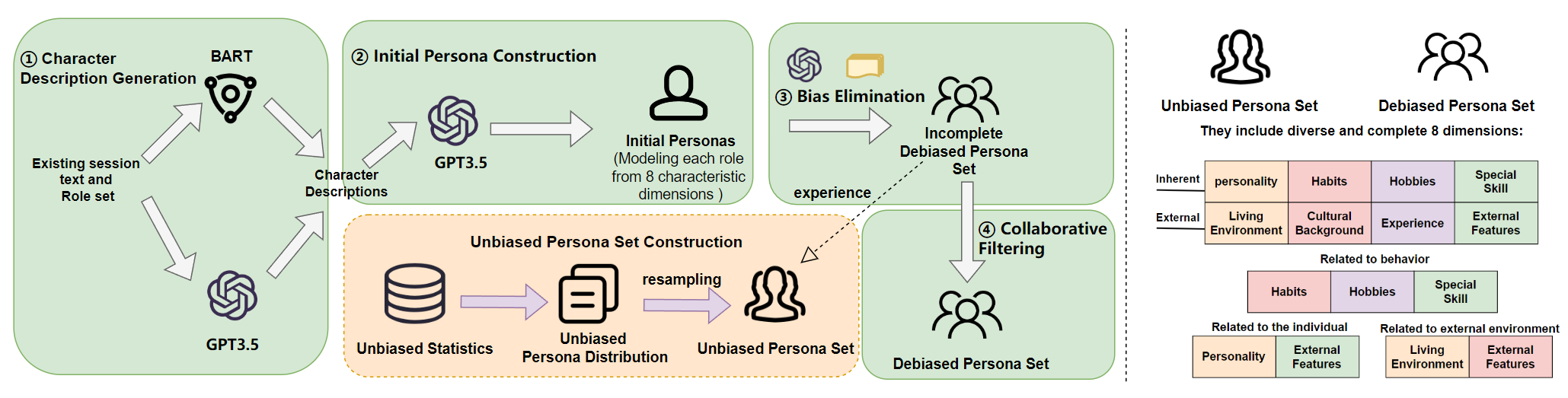}
    \caption{The UPCS framework innovatively combines debiased and unbiased personas for personalized dialogue generation, with an eight-dimensional role modeling for general personality.}
    \vspace{-14pt}
    \label{fig:upcs}
\end{figure*}

Toxic biases in Personas can alienate and frustrate users during interactions, contrary to the original intention of enhancing user experience, and even raise severe social and ethical concerns such as \cite{Wan2023Are,lu2023large}. Hence, addressing bias in Persona design is of paramount importance.  To tackle this, we propose a novel framework, \textbf{U}nbiased \textbf{P}ersona \textbf{C}onstruction \textbf{S}ystem (\textbf{UPCS}), which integrates explicit bias elimination mechanisms with multi-dimensional character construction in dialogue system character descriptions, aims to obtain unbias and reliable Personas for dialogue generation.

To be specific, UPCS constructs two persona sets designed to minimize biases. By integrating the automated functionalities of ChatGPT with manual script validation, UCPS effectively mitigates harmful biases during persona construction, resulting in the \textbf{Debiased Persona Set}. Additionally, UCPS extracts unbiased distributions of personas from real-world statistics and constructs the \textbf{Unbiased Persona Set} by sampling according to these distributions. Both the Debiased Persona Set and the Unbiased Persona Set are utilized simultaneously in subsequent personalized dialogue generation.
We conducted a thorough evaluation of the UPCS framework and existing Persona construction methods using auto-mated language tools (ChatGPT) and 50 human evaluators. The results indicate that both language automation tools and human evaluators consistently agree that the proposed UPCS framework significantly reduces harmful biases and surpasses traditional methods in terms of role description diversity and user satisfaction.

\section{Related Work}
\label{sec:relwork}

\noindent{\textbf{Persona Construction}}
The construction of personas involves various methodologies designed to create detailed user profiles through diverse data sources and analytical techniques. One prominent approach is integrating the Knowledge Graph Construction Methods \cite{zou2020survey,chen2020knowledge,wei2024kicgpt} with persona construction, which can leverage interconnected information in knowledge graphs to develop comprehensive user personas \cite{Huang2021Knowledge}. For instance, PeaCok \cite{gao2023peacokpersonacommonsenseknowledge} utilizes knowledge graphs to enhance personas, enabling them to possess a broader range of knowledge and thereby making conversations more coherent and contextually relevant. Another technique, the Direct Extraction Method, builds user personas by directly extracting features from users' social media activity, browsing history, and other data sources \cite{Salminen2019Automatic, Jung2017Persona}. Additionally, the Automated Generation Method employs natural language processing and machine learning techniques to automatically generate user personas, proving particularly effective for applications involving large and complex data structures \cite{10.1007/978-3-031-48057-7_14, Anik2022Machine}.

\noindent{\textbf{Bias Mitigation}}
Researchers have developed various techniques to mitigate biases in machine learning models across diverse domains. In data preprocessing, resampling methods, such as oversampling or undersampling, have been employed to address class imbalances, which helps in reducing biases inherent in imbalanced datasets \cite{Kim2023Optimal, Lin2016Clustering-based}. In clustering, methods such as redefined cluster centers have been proposed to address inherent data biases \cite{Gan2023Fair, Bercea2018On}. In image recognition, adversarial training techniques have been employed to reduce racial and gender biases \cite{Dhar2020An, Wang2018Balanced}. For text classification, specialized loss functions have been introduced to significantly mitigate language biases \cite{Wang2023Research, Qian2019Reducing}. Additionally, in recommendation systems, debiasing efforts aim to balance exposure rates among different groups, thereby addressing group-based biases within these algorithms \cite{10.1145/3564284}.

\section{Methodology}
\label{sec:frame}

As illustrated in Fig. \ref{fig:upcs}. the UCPS framework constructs two persona sets aimed at minimizing biases: the Debiased Persona Set and the Unbiased Persona Set. These sets are then employed in tandem to generate personalized dialogues.

\subsection{Debiased Persona Set Construction}

As illustrated in Fig. \ref{fig:upcs}, the green blocks represent the construction of the Debiased Persona Set, which involves four phases detailed below.

\subsubsection{Phase 1: Character Description Generation}
We employ BART \cite{10.1007/978-3-030-50334-5_7} to generate personas' motivations, abilities, desires, and other defining characteristics, and GPT-3.5 \cite{openai2023gpt} to create simple descriptions of the personas. These attributes collectively form the \textbf{Character Description}, providing essential and detailed information that underpins the subsequent development of their personas.

\subsubsection{Phase 2: Initial Persona Construction}
\begin{figure}[t]
\centering
\includegraphics[width=1.0\columnwidth]{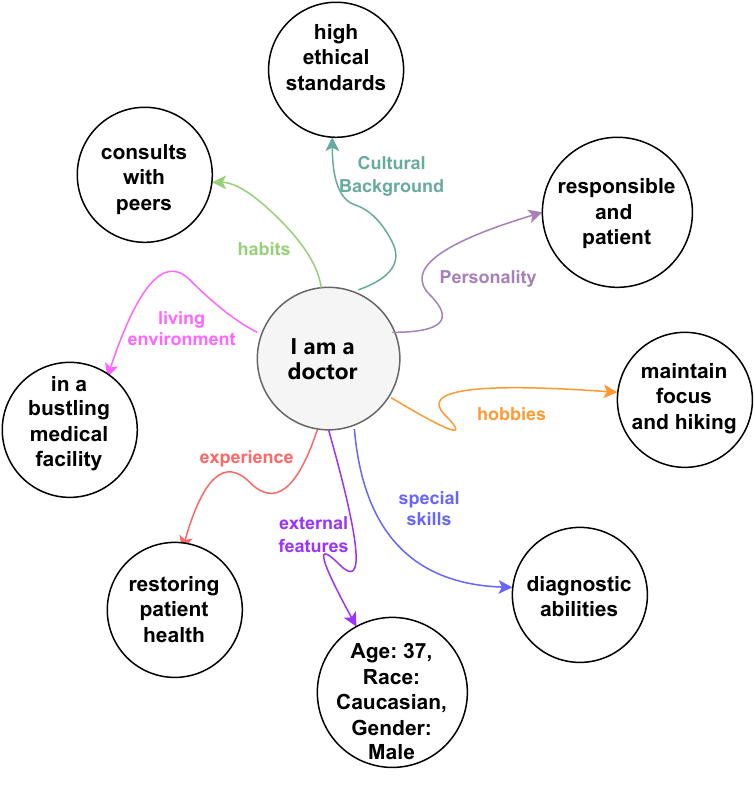}
\caption{Rough descriptions of the eight dimensions of character roles.}
\vspace{-10pt}
\label{fig:persona_illustrate}
\end{figure}

In this phase, the detailed Character Descriptions generated in the preliminary stage are input into GPT-3.5 \cite{openai2023gpt} to construct \textbf{Initial Personas}. These personas encompass comprehensive eight diverse dimensions to model a role: personality traits, experiences, interests and hobbies, special skills, living environment, habits, cultural background, and external characteristics. These dimensions can be further categorized into different groups according to psychological taxonomy, as illustrated in the right part of Fig \ref{fig:upcs}. Additionally, we provide an example of a persona with these dimensions in Fig \ref{fig:persona_illustrate}. The explanation of these dimensions is as follows: \\

\noindent{\textbf{Personality:}} Personality encompasses the intrinsic traits or qualities of an individual, such as cheerfulness, empathy, or high levels of organization.

\noindent{\textbf{Experience:}} Experience pertains to the extrinsic events or activities in which an individual has participated, including professional roles and volunteer programs \cite{BICKHARD1994229}.

\noindent{\textbf{Hobbies:}} Hobbies refer to leisure activities or interests, such as painting, playing basketball, or stamp collecting \cite{doi:10.1080/00220973.1940.11010173}.

\noindent{\textbf{Special Skills:}} Special skills denote unique talents or abilities, such as multilingual fluency or exceptional culinary expertise.

\noindent{\textbf{Living Environment:}}Living environment describes the settings in which an individual resides, such as a city apartment, countryside farmhouse, or metropolitan area \cite{Ayoub2020}.

\noindent{\textbf{Habits:}} Habits refer to regular behaviors or routines, such as morning runs or bedtime reading \cite{McCloskey2021You}.

\noindent{\textbf{Cultural Background:}} Cultural background includes an individual's ethnicity, religion, and language, for instance, being Chinese-American and practicing Christianity. \cite{Muskhanova2021Psychological}

\noindent{\textbf{External Features (Age, Race, Gender):}} External features describe physical attributes, including age, race, and gender, such as being a 30-year-old Asian woman or a 50-year-old white man.

\subsubsection{Phase 3: Bias Elimination}
\label{bias elimination}
To eliminate biases in the initial personas, all sentences within these personas were systematically reviewed using an automated tool, specifically GPT-3.5. This tool was tasked with identifying biases within the personas, locating the corresponding biased sentences, and deleting them. Following this automated bias elimination, we performed a subsequent validation using a script that includes common bias expressions manually collected by us. This script calculates the BM25 text similarity score between these biased expressions and the remaining parts of the personas. If the similarity score exceeds a threshold of 0.75, the sentences are re-evaluated by GPT-3.5 to verify and potentially remove any remaining biases.

However, it is important to note that the initial personas do not contain complete information across all eight dimensions for each persona, and the bias elimination process exacerbates this incompleteness.

\subsubsection{Phase 4: Collaborative Filtering}

Collaborative filtering \cite{10.1145/371920.372071} is a widely used technique in recommendation systems that generates recommendations for users with sparse data by leveraging the preferences of similar, well-informed users. In this Phase, we apply the same idea as collaborative filtering to complete missing dimensions in personas with the dimensions in similar personas. Specifically, we first calculate the similarity scores among each pair of personas, which is formulated as a sum of cosine similarity \cite{Kryszkiewicz2014The} and the Pearson correlation coefficient:

$$
S(P_i, P_j) = \alpha \cdot \text{CS}(P_i, P_j) + \beta \cdot \text{PC}(P_i, P_j)
$$

where \(\text{CS}(P_i, P_j)\) and \(\text{PC}(P_i, P_j)\) are defined as follows:

$$
\text{CS}(P_i, P_j) = \frac{E_{P_i} \cdot E_{P_j}}{\|E_{P_i}\| \|E_{P_j}\|}
$$

$$
\text{PC}(P_i, P_j) = \frac{\sum_{k=1}^n (C_{ik} - \bar{E_{P_i}})(E_{P_ik} - \bar{E_{P_j}})}{\sqrt{\sum_{k=1}^n (E_{P_ik} - \bar{E_{P_i}})^2} \sqrt{\sum_{k=1}^n (E_{P_jk} - \bar{E_{P_j}})^2}}
$$

Here, \( S(P_i, P_j) \) represents the similarity score between personas \( P_i \) and \( P_j \), \( E_{P_i}\) and \( E_{P_j}\) are the text embedding of \(P_i\) and \(P_j\) via pre-trained BERT \cite{devlin2019bertpretrainingdeepbidirectional}, \(E_{P_ik}\) and \(E_{P_jk}\) are the \(k\)-th elements of \(E_{P_i}\) and \(E_{P_j}\). \(\bar{E_{P_i}}\) and \(\bar{E_{P_j}}\) are the means of \(E_{P_i}\) and \(E_{P_j}\), respectively.

Next, for each persona \(P_m\) with missing dimensions, we fill its missing dimensions using values from its most similar persona \(P_n = \arg\max_{P_i} S(P_m, P_i)\). To ensure that the collaborative filtering is based on sufficiently high similarity and to prevent unreasonable filling, we perform the filling operation only if the BM25 text similarity score \cite{Robertson2009The} \(BM25(P_m, P_n) \geq \theta\), where the threshold \(\theta\) is set to 0.5 in our implementation.  Consequently, after this stage, the Debiased Persona Set is completed and prepared for further use. 

\subsection{Unbiased Persona Set Construction}
The Unbiased Persona Set is constructed by resampling the dimensions, excluding experience, within the incomplete Debiased Persona Set (generated after Phase 3 of the Debiased Persona Set construction in Sec. \ref{bias elimination}), based on a predefined unbiased distribution \(D_{unbias}\). Consequently, the resampled Unbiased Persona Set maintains the same scale as the Incomplete Debiased Persona Set. This method ensures that the resulting personas inherently exhibit unbiased characteristics, thereby mitigating the unbalanced distribution of personas and particularly enhancing the representation of long-tail personas, which often correspond to marginalized groups more susceptible to biases. 

To be Specific, the predefined unbiased distribution \(D_{unbias}\) is derived from a collection of unbiased real-world data, such as global population statistics on age, race, and gender provided by the World Health Organization \cite{owid-age-structure}. These statistics inform the default distribution for dimensions like External Features, ensuring a more balanced and representative Unbiased Persona Set.
Similarly, \(D_{unbias}\) prescribes default values and distributions for all attributes across the seven dimensions, excluding the experience dimension. The resampling of values for all attributes is performed according to these distributions, ensuring a more unbiased representation. The experience dimension, however, remains unchanged to preserve consistency in context and background between the Unbiased Persona Set and the original set. Although \(D_{unbias}\) is designed to reflect objective statistical data, these data are subject to change over time. Therefore, the Unbiased Persona Set incorporates a flexible interface that allows for the input of custom attributes and distributions, \(D_{custom}\), enabling the resampling of dimension attributes in the Unbiased Persona Set according to \(D_{custom}\).

\begin{table*}[h] 
\caption{Pairwise Comparison of UPCS and Baselines on Subjective Human Evaluation Metrics and GPT-3.5 Evaluated Bias Quantity.}
\centering
\resizebox{\linewidth}{!}{%
\begin{tabular}{lccccccc}
\toprule
Model & Bias & \multicolumn{1}{c}{Fluency} & \multicolumn{1}{c}{Emotional Expression} & \multicolumn{1}{c}{Engagement} & \multicolumn{1}{c|}{Personality Expression} & Bias \\
& \multicolumn{1}{c}{win (\%)} & \multicolumn{1}{c}{win (\%)} & \multicolumn{1}{c}{win (\%)} & \multicolumn{1}{c}{win (\%)} & 
\multicolumn{1}{c|}{win (\%)} &
\multicolumn{1}{c}{Quantity}\\
\midrule
\textsc{P$^{2}$Bot} + UPCS  & 55 & 70 & 65 & 65 & 75 & 906 \\
\textsc{P$^{2}$Bot} & 45 & 30 & 35 & 35 & 25 & 1075 \\
\midrule
\textsc{P$^{2}$Bot} + UPCS  & 55 & 60 & 65 & 45 & 50 & 859 \\
\textsc{P$^{2}$Bot} + PeaCok & 45 & 40 & 35 & 55 & 50 & 1119 \\
\bottomrule
\end{tabular}
}
\label{table1} 
\end{table*}

\begin{table}[t]
\caption{Comparison of UPCS and Baselines on Objective Hard-computed Metrics.}
\centering
\resizebox{\linewidth}{!}{%
\begin{tabular}{lccccc}
\toprule
Model & Hits@1 (\%) & F1 (\%) & BLEU (\%) & TB rank & UTR rank \\
\midrule
\textsc{P$^{2}$Bot} + UPCS       & 67.0 & 18.37 & 0.68 & \textbf{964.14} & \textbf{986.60}\\
\textsc{P$^{2}$Bot} + PeaCok & 69.7 & \textbf{18.39} & \textbf{0.73} & 992.44  & 993.01\\
\textsc{P$^{2}$Bot}            & \textbf{69.8} & 18.29 & 0.66 & 1,016.91 & 993.88\\
\bottomrule
\end{tabular}
}
\label{table2}  
\end{table}

% Table 3
\begin{table}[t]
\caption{UPCS Ablations on Objective Hard-computed Metrics}
\centering
\resizebox{\linewidth}{!}{%
\begin{tabular}{lccccc}
\toprule
Model & Hits@1 (\%) & F1 (\%) & BLEU (\%) & TB rank & UTR rank\\
\midrule
\textsc{P$^{2}$Bot} + UPCS       & 67.0 & \textbf{18.37} & 0.68 & \textbf{962.04} & \textbf{976.85}\\
\textsc{P$^{2}$Bot} + Unbiased & 65.0 & 18.26 & \textbf{0.76} & 991.64 & 988.00\\
\textsc{P$^{2}$Bot} + Debiased  & 67.0 & 18.14 & 0.63 & 969.66 & 988.48\\
\textsc{P$^{2}$Bot}            & \textbf{69.8} & 18.29 & 0.66 & 1,014.33 & 984.35\\
\bottomrule
\end{tabular}
}
\label{table3}  
\end{table}

\section{Experiment}

\subsection{Experiment Setup}
\noindent{\textbf{Implementation Details}} 
Following the settings in the knowledge-enhanced Persona construction framework Peacok \cite{gao2023peacokpersonacommonsenseknowledge}, we leveraged the conversations provided in the ConvAI2 \cite{dinan2019secondconversationalintelligencechallenge} PERSONA-CHAT \cite{zhang2018personalizingdialogueagentsi} dataset and our UPCS generated Personas to train a \textsc{P$^{2}$Bot} model \cite{liu-etal-2020-impress} to generate persona-based personalized dialogues. To be specific, the ConvAI2 PERSONA-CHAT dataset consists of 17,878 dialogue segments, where each dialogue segment includes detailed conversation text and two personas containing separate descriptions about the two persons performing this conversation. We use these conversation texts to train \textsc{P$^{2}$Bot}, but substitute the intrinsic personas in the dataset to our UPCS generated Persona Sets (both Debiased and Unbiased Sets) during training. 

\noindent{\textbf{Baselines}}
Our baseline models encompass Personalized dialogues generated by \textsc{P$^{2}$Bot} fine-tuned using: 1) The intrinsic personas extracted from the dataset.
2) Knowledge-augmented Personas constructed by Peacok \cite{gao2023peacokpersonacommonsenseknowledge}.

\noindent{\textbf{Evaluation Metrics}}
To showcase the effectiveness of our UPCS compared to these baselines, we conduct a comprehensive evaluation encompassing both traditional quality metrics for dialogue generation and the detection of toxic biases within the generated dialogues. 

The evaluation of dialogue quality involves objective hard-computed measures Hits@1, F1 Score, and BLEU Score \cite{papineni2002bleu}, as well as subjective human evaluation on fluency, engagement, emotional expression, and personality expression. The subjective human evaluations are carried out by 50 evaluators hired through Taobao, where the evaluators are required to determine the superior model in pairwise competitions.

For bias evaluation, we report two objective metrics: "TB rank" and "UTR rank". They are the average ranking number of dialogue sentences according to bias (with higher rankings indicating less bias), determined using the specialized bias-detecting models Toxic-BERT \cite{Detoxify} and Unbiased-Toxic-RoBERTa \cite{Detoxify}. Additionally, we report a metric called "Bias Quantity," computed by GPT-3.5, which indicates the number of sentences identified as relatively biased compared to the other model in pairwise competitions. We also include results from a human-evaluated bias competition, conducted similarly to the subjective human evaluations used for assessing dialogue quality.

\section{Results}
\label{sec:data}

\subsection{Evaluation Results}
Table \ref{table1} presents the subjective comparison results (pairwise competition) between the proposed UPCS and the Baselines. The results show that, compared to using the original Persona from the dataset, the Unbiased and Debiased Persona Sets constructed using UPCS exhibit superior performance in generating personalized dialogues across all five human evaluation metrics: Fluency, Emotional Expression, Engagement, Personality Expression, and Bias, as well as the Bias Quantity metric evaluated by GPT. This indicates that the Personas generated by UPCS can effectively reduce bias in dialogues while significantly enhancing dialogue quality. When compared with the state-of-the-art Persona construction method, Peacock, the Debiased and Unbiased Personas generated by UPCS still perform better in five out of six metrics (except Engagement). Therefore, in the subjective evaluations by humans and GPT-3.5, UPCS achieves state-of-the-art performance in both personalized dialogue quality and bias reduction.

Table \ref{table2} showcases the objective comparison metrics between UPCS and the Baseline. The observations reveal that, compared to the Baselines, UPCS shows significant advantages in TB rank and UTR rank scores, demonstrating its effectiveness in reducing bias in dialogues. Additionally, UPCS exhibits comparable performance in Hits@1, F1, and BLEU scores, indicating that this bias reduction strategy does not compromise dialogue quality. Therefore, whether from the subjective evaluations by humans and GPT-3.5 or from objective metrics, UPCS consistently proves its effectiveness in bias reduction and its stability in maintaining dialogue quality.

\subsection{Ablation Study}
To validate the effectiveness of UPCS, we conducted an ablation study as shown in Table \ref{table3}, which compares the performance of using the complete UPCS (\textsc{P$^{2}$Bot}+UPCS), only the Unbiased Persona Set (\textsc{P$^{2}$Bot}+Unbiased), only the Debiased Persona Set (\textsc{P$^{2}$Bot}+Debiased), and not using UPCS at all (\textsc{P$^{2}$Bot}) on objective hard-computed metrics. The results indicate that the complete use of UPCS, which combines both the Debiased and Unbiased Persona Sets, achieves the best performance in reducing bias, as demonstrated by their superior TB rank and UTR rank. Removing either Debiased or Unbiased Persona set can weaken the bias mitigation effect, but the performance still surpasses that of not using them. This supports our hypothesis that the Debiased Set and Unbiased Set independently contribute to reducing bias in different ways: the Debiased Set improves bias in persona expression, while the Unbiased Set corrects bias in the original persona distribution, thus enhancing the expression of marginalized groups. For other metrics, using either the Debiased or Unbiased Persona Set alone remains comparable or even superior.

\section{Conclusion}
In this paper, we have presented a novel persona construction framework, UPCS, which aims to alleviate toxic bias within personalized dialogue generation. Through thorough comparisons using both objective hard-computed metrics and subjective human evaluations, we have demonstrated that the proposed UPCS framework significantly reduces biases in personalized dialogues compared to state-of-the-art persona construction methods, while maintaining comparable dialogue quality. In our future work, we will investigate the UPCS framework model combined with debiased external knowledge and explore the potential of the model in downstream tasks, such as personalized content recommendation.

\section{Limitations and Future Directions}

First, the eight dimensions of personality construction in UPCS are designed for general personality traits. However, the real world may involve more complex role modeling requirements, especially for dialogue generation under special topic constraints (e.g., healthcare, finance, entertainment). In these cases, while the bias mitigation approach provided by UPCS still holds value, it is essential to combine it with persona attributes specifically designed for the topic's context and their corresponding unbiased distributions. We leave these topic-constrained debiasing tasks for future research.

Secondly, the sustainability of bias elimination systems poses a challenge due to biases being intrinsically linked to dynamic and evolving universal values, which can vary across different times and contexts. To align with emerging mainstream values, the \(D_{unbias}\) component in UPCS can be substituted through offline updates. More advanced future research could focus on exploring online bias detection and integrating dynamic adjustments during dialogue generation to enhance the system's adaptability to evolving values. 

Third, while UPCS primarily focuses on eliminating biases inherent in the data and balancing the inherent distribution imbalances within the data, another compatible approach is to incorporate bias mitigation directly into model training. For instance, adding a penalty term related to bias in the loss function of dialogue generation models could be a valuable exploration direction for the future.

Lastly, the integration of graph algorithms and large language models \cite{wu2024backdoorgraphcondensation,10102293}, as well as the fusion of multimodal \cite{chen2024large} large language model methods in UPCS, is deserving of further exploration.
\bibliographystyle{IEEEtran}
\bibliography{paper_arxiv}

\end{document}